\documentclass{MITcsail}
\usepackage{times}
\usepackage{calc}
\usepackage{xlop}
\usepackage{amsmath}
\usepackage{amsthm}  
\usepackage{amssymb}
\usepackage{bm}
\usepackage{bbm}
\usepackage{color}
\usepackage{mathtools} 
\usepackage{algorithm,algorithmicx,algpseudocode}

\title{Liquid Structural State-Space Models}

\author
{Ramin Hasani~*~\footnote{Code Repository: \texttt{https://github.com/raminmh/liquid-s4}}, Mathias Lechner~*, Tsun-Hsuan Wang,~Makram Chahine,~Alexander Amini, Daniela Rus\\
\vspace{1em} % Space between authors and afilliations
\normalfont{\small 
Computer Science and Artificial Intelligence Lab (CSAIL) \\
Massachusetts Institute of Technology (MIT)\\
Cambridge, 02139, MA} \\
* indicates authors with equal contributions \\
Correspondence to \texttt{rhasani@mit.edu}
\vspace{2em}
}

\begin{document}

\maketitle
\thispagestyle{firstpagestyle} % Draws the header on the first page

\begin{abstract}
A proper parametrization of state transition matrices of linear state-space models (SSMs) followed by standard nonlinearities enables them to efficiently learn representations from sequential data, establishing the state-of-the-art on a large series of long-range sequence modeling benchmarks. In this paper, we show that we can improve further when the structural SSM such as S4 is given by a linear liquid time-constant (LTC) state-space model. LTC neural networks are causal continuous-time neural networks with an input-dependent state transition module, which makes them learn to adapt to incoming inputs at inference. We show that by using a diagonal plus low-rank decomposition of the state transition matrix introduced in S4, and a few simplifications, the LTC-based structural state-space model, dubbed Liquid-S4, achieves the new state-of-the-art generalization across sequence modeling tasks with long-term dependencies such as image, text, audio, and medical time-series, with an average performance of 87.32\% on the Long-Range Arena benchmark. On the full raw Speech Command recognition dataset Liquid-S4 achieves 96.78\% accuracy with 30\% reduction in parameter counts compared to S4. The additional gain in performance is the direct result of the Liquid-S4's kernel structure that takes into account the similarities of the input sequence samples during training and inference.
\end{abstract}

\section{Introduction}

Learning representations from sequences of data requires expressive temporal and structural credit assignment. In this space, the continuous-time neural network class of liquid time-constant networks (LTC) \citep{hasani2021liquid} has shown theoretical and empirical evidence for their expressivity and their ability to capture the cause and effect of a given task from high-dimensional sequential demonstrations \citep{lechner2020neural,vorbach2021causal}. Liquid networks are nonlinear state-space models (SSMs) with an input-dependent state transition module that enables them to learn to adapt the dynamics of the model to incoming inputs, at inference, as they are dynamic causal models \citep{friston2003dynamic}. Their complexity, however, is bottlenecked by their differential equation numerical solver that limits their scalability to longer-term sequences. How can we take advantage of LTC's generalization and causality capabilities and scale them to competitively learn long-range sequences without gradient issues, compared to advanced recurrent neural networks (RNNs) \citep{rusch2021coupled,erichson2021lipschitz,gu2020hippo}, convolutional neural networks (CNNs) \citep{lea2016temporal,romero2021ckconv,cheng2022classification}, and attention-based models \citep{vaswani2017attention}? 

In this work, we set out to leverage the elegant formulation of structural state-space models (S4) \citep{gu2022efficiently} to obtain linear liquid network instances that possess the approximation capabilities of both S4 and LTCs. This is because structural SSMs are shown to largely dominate advanced RNNs, CNNs, and Transformers across many data modalities such as text, sequence of pixels, audio, and time series \citep{gu2021combining,gu2022efficiently,gu2022parameterization,gupta2022diagonal}. Structural SSMs achieve such impressive performance by using three main mechanisms: 1) High-order polynomial projection operators (HiPPO) \citep{gu2020hippo} that are applied to state and input transition matrices to memorize signals' history, 2) diagonal plus low-rank parametrization of the obtained HiPPO \citep{gu2022efficiently}, and 3) an efficient (convolution) kernel computation of an SSM's transition matrices in the frequency domain, transformed back in time via an inverse Fourier transformation \citep{gu2022efficiently}. 

To combine S4 and LTCs, instead of modeling sequences by linear state-space models of the form $\dot{x} = \textbf{A}~x + \textbf{B}~u$, $y = \textbf{C}~x$, (as done in structural and diagonal SSMs \citep{gu2022efficiently,gu2022parameterization}), we propose to use a linearized LTC state-space model \citep{hasani2021liquid}, given by the following dynamics: $ \dot{x} = (\textbf{A} + \textbf{B}~u)~x + \textbf{B}~u$, $y= \textbf{C}~x$. We show that this dynamical system can also be efficiently solved via the same parametrization of S4, giving rise to an additional convolutional Kernel that accounts for the similarities of lagged signals. We call the obtained model Liquid-S4. Through extensive empirical evaluation, we show that Liquid-S4 consistently leads to better generalization performance compared to all variants of S4, CNNs, RNNs, and Transformers across many time-series modeling tasks. In particular, we achieve SOTA performance on the Long Range Arena benchmark \citep{tay2020long} with an average of 87.32\%. To sum up, we make the following contributions:
\begin{enumerate}
\setlength{\itemsep}{0pt}
\setlength{\parskip}{0pt}
    \item We introduce Liquid-S4, a new state-space model that encapsulates the generalization and causality capabilities of liquid networks as well as the memorization, efficiency and scalability of S4.
    \item We achieve State-of-the-art performance on pixel-level sequence classification, text, speech recognition and all six tasks of the long-range arena benchmark with an average accuracy of 87.32\%. On the full raw Speech Command recognition dataset Liquid-S4 achieves 96.78\% accuracy with 30\% reduction in parameter. Finally on the BIDMC vital signs dataset Liquid-S4 achieves SOTA in all modes.
\end{enumerate}

\section{Related Works}

\textbf{Learning Long-Range Dependencies with RNNs.} 
Sequence modeling can be performed autoregressively with RNNs which possess persistent states \citep{little1974existence} originated from Ising \citep{brush1967history} and Hopfield networks \citep{hopfield1982neural,ramsauer2020hopfield}. Discrete RNNs approximate continuous dynamics step-by-steps via dependencies on the history of their hidden states,  and continuous-time (CT) RNNs use ordinary differential equation (ODE) solvers to unroll their dynamics with more elaborate temporal steps \citep{funahashi1993approximation}. 

CT-RNNs can perform remarkable credit assignment in sequence modeling problems both on regularly sampled, irregularly-sampled data \citep{pearson2003imbalanced,li2016scalable,belletti2016scalable,roy2020robust,foster1996wavelets,amigo2012transcripts,kowal2019functional}, by turning the spatiotemproal dependencies into vector fields \citep{chen2018neural}, enabling better generalization and expressivity \citep{massaroli2020dissecting,hasani2021liquid}. Numerous works have studied their characteristics to understand their applicability and limitations in learning sequential data and flows \citep{lechner2019designing,dupont2019augmented,durkan2019neural,jia2019neural,grunbacher2021verification,hanshu2020robustness,holl2020learning,quaglino2020snode,kidger2020neural,hasani2020natural,liebenwein2021sparse,gruenbacher2022gotube}.

However, when these RNNs are trained by gradient descent \citep{rumelhart1986learning,allen2019can,sherstinsky2020fundamentals}, they suffer from the vanishing/exploding gradients problem, which makes difficult the learning of long-term dependencies in sequences \citep{hochreiter1991untersuchungen,bengio1994learning}. This issue happens in both discrete RNNs such as GRU-D with its continuous delay mechanism \citep{che2018recurrent} and Phased-LSTMs \citep{neil2016phased}, and continuous RNNs such as ODE-RNNs \citep{rubanova2019latent}, GRU-ODE \citep{de2019gru}, Log-ODE methods \citep{morrill2020neural} which compresses the input time-series by time-continuous path signatures \citep{friz2010multidimensional}, and neural controlled differential equations  \citep{kidger2020neural}, and liquid time-constant networks (LTCs) \citep{hasani2021liquid}. 

Numerous solutions have been proposed to resolve these gradient issues to enable long-range dependency learning. Examples include discrete gating mechanisms in LSTMs \citep{hochreiter1997long,greff2016lstm,hasani2019response}, GRUs \citep{chung2014empirical}, continuous gating mechanisms such as CfCs \citep{hasani2021closed},  hawks LSTMs \citep{mei2017neural}, IndRNNs \citep{li2018independently}, state regularization \citep{wang2019state}, unitary RNNs \citep{jing2019gated}, dilated RNNs \citep{chang2017dilated}, long memory stochastic processes \citep{greaves2019statistical}, recurrent kernel networks \citep{chen2019recurrent}, Lipschitz RNNs \citep{erichson2021lipschitz}, symmetric skew decomposition \citep{wisdom2016full}, infinitely many updates in iRNNs \citep{kag2019rnns}, coupled oscillatory RNNs (coRNNs) \citep{rusch2021coupled}, mixed-memory RNNs \citep{lechner2021mixed}, and Legendre Memory Units \citep{voelker2019legendre}.

\noindent \textbf{Learning Long-range Dependencies with CNNs and Transformers.} RNNs are not the only solution to learning long-range dependencies. Continuous convolutional kernels such as CKConv \citep{romero2021ckconv} and \citep{romero2021flexconv}, and circular dilated CNNs \citep{cheng2022classification} have shown to be efficient in modeling long sequences faster than RNNs. There has also been a large series of works showing the effectiveness of attention-based methods for modeling spatiotemporal data. A large list of these models is listed in Table \ref{tab:lra}. These baselines have recently been largely outperformed by the structural state-space models \citep{gu2022efficiently}.

\noindent \textbf{State-Space Models.} SSMs are well-established frameworks to study deterministic and stochastic dynamical systems \citep{kalman1960new}. Their state and input transition matrices can be directly learned by gradient descent to model sequences of observations \citep{lechner2020gershgorin,hasani2021liquid,gu2021combining}. In a seminal work, \citet{gu2022efficiently} showed that with a couple of fundamental algorithmic methods on memorization and computation of input sequences, SSMs can turn into the most powerful sequence modeling framework to-date, outperforming advanced RNNs, temporal and continuous CNNs \citep{cheng2022classification,romero2021ckconv,romero2021flexconv} and a wide variety of Transformers \citep{vaswani2017attention}, available in Table \ref{tab:lra} by a significant margin. 

The key to their success is their diagonal plus-low rank parameterization of the transition matrix of SSMs via higher-order polynomial projection (HiPPO) matrix \citep{gu2020hippo} obtained by a scaled Legendre measure (LegS) inspired by the Legendre Memory Units \citep{voelker2019legendre} to memorize input sequences, a learnable input transition matrix, and an efficient Cauchy Kernel algorithm, results in obtaining structural SSMs named S4. It was also shown recently that diagonal SSMs (S4D) \citep{gupta2022diagonal} could be as performant as S4 in learning long sequences when parametrized and initialized properly \citep{gu2022parameterization,gu2022train}. There was also a new variant of S4 introduced as simplified-S4 (S5) \cite{smith2022simplified} that tensorizes the 1-D operations of S4 to gain a more straightforward realization of SSMs. Here, we introduce Liquid-S4, which is obtained by a more expressive SSM, namely liquid time-constant (LTC) representation \citep{hasani2021liquid} which achieves SOTA performance across many benchmarks.

\section{Setup and Methodology}
In this section, we first revisit the necessary background to formulate our Liquid Structural State-Space Models. We then set up and sketch our technical contributions. 

\subsection{Background}
We aim to design an end-to-end sequence modeling framework built by SSMs. A continuous-time SSM representation of a linear dynamical system is given by the following set of equations: 

\begingroup
\small
\begin{align}
\label{eq:ss_continuous}
    \dot{x}(t)  = \textbf{A}~x(t) + \textbf{B}~u(t),~~~~y(t) = \textbf{C}~x(t) + \textbf{D}~u(t).
\end{align}
\endgroup

Here, $x(t)$ is an $N$-dimensional latent state, receiving a 1-dimensional input signal $u(t)$, and computing a 1-dimensional output signal $y(t)$. $\textbf{A}^{(N \times N)}$, $\textbf{B}^{(N \times 1)}$, $\textbf{C}^{(1 \times N)}$ and $\textbf{D}^{(1 \times 1)}$ are system's parameters. For the sake of brevity, throughout our analysis, we set $\textbf{D} = 0$ as it can be added eventually after construction of our main results in the form of a skip connection \citep{gu2022efficiently}. 

\noindent \textbf{Discretization of SSMs.} In order to create a sequence-to-sequence model similar to a recurrent neural network (RNN), we discretize the continuous-time representation of SSMs by the trapezoidal rule (bilinear transform)\footnote{$s \leftarrow \frac{2}{\delta t} \frac{1-z^{-1}}{1+z^{-1}}$} as follows (sampling step = $\delta t$) \citep{gu2022efficiently}:
\begingroup
\small
\begin{align}
\label{eq:ss_disc}
    x_k = \overline{\textbf{A}}~x_{k-1} + \overline{\textbf{B}}~u_k,~~~~y_k = \overline{\textbf{C}}~x_k \\ \nonumber
\end{align}
\endgroup

This is obtained via the following modifications to the transition matrices:

\begingroup
\small
\begin{align}
\label{eq:ss_disc_params}
    \overline{\textbf{A}} = (\textbf{I} - \frac{\delta t}{2} \textbf{A})^{-1} (\textbf{I} + \frac{\delta t}{2} \textbf{A}),~~~~~
    \overline{\textbf{B}} = (\textbf{I} - \frac{\delta t}{2} \textbf{A})^{-1}~\delta t~\textbf{B},~~~~\overline{\textbf{C}} = \textbf{C} 
\end{align}
\endgroup

With this transformation, we constructed a discretized seq-2-seq model that can map the input $u_k$ to output $y_k$, via the \emph{hidden} state $x_k \in \mathbbm{R}^N$. $\overline{\textbf{A}}$ is the hidden transition matrix, $\overline{\textbf{B}}$ and $\overline{\textbf{C}}$ are input and output transition matrices, respectively.

\noindent \textbf{Creating a Convolutional Representation of SSMs.} The system described by (\ref{eq:ss_disc}) and (\ref{eq:ss_disc_params}), can be trained via gradient descent to learn to model sequences, in a sequential manner which is not scalable. To improve this, we can write the discretized SSM in (\ref{eq:ss_disc}) as a discrete convolutional kernel. To construct the convolutional kernel, let us unroll the system (\ref{eq:ss_disc}) in time as follows, assuming a zero initial hidden states $x_{-1} = 0$:

\begingroup
\small
\begin{align}
    \label{eq:unroll_ssm}
     x_0 &= \overline{\textbf{B}} u_0,~~~~~~~x_1= \overline{\textbf{A}}\overline{\textbf{B}} u_0 + \overline{\textbf{B}} u_1,~~~~~~~~~x_2 = \overline{\textbf{A}}^2\overline{\textbf{B}} u_0 + \overline{\textbf{A}}\overline{\textbf{B}} u_1 + \overline{\textbf{B}} u_2,~~~~~~~~~~~~\dots\\ \nonumber
     y_0 &= \overline{\textbf{C}}\overline{\textbf{B}} u_0,~~~~y_1= \overline{\textbf{C}}\overline{\textbf{A}}\overline{\textbf{B}} u_0 + \overline{\textbf{C}}\overline{\textbf{B}} u_1,~~~~y_2 = \overline{\textbf{C}}\overline{\textbf{A}}^2\overline{\textbf{B}} u_0 + \overline{\textbf{C}}\overline{\textbf{A}}\overline{\textbf{B}} u_1 + \overline{\textbf{C}}\overline{\textbf{B}} u_2,~~~~\dots
\end{align}
\endgroup

The mapping $u_k \rightarrow y_k$ can now can be formulated into a convolutional kernel explicitly:

\begingroup
\small
\begin{align}
\label{eq:conv_kernel}
    y_k = \overline{\textbf{C}}\overline{\textbf{A}}^k\overline{\textbf{B}} u_0 + \overline{\textbf{C}}\overline{\textbf{A}}^{k-1}\overline{\textbf{B}} u_1 + \dots \overline{\textbf{C}}\overline{\textbf{A}}\overline{\textbf{B}} u_{k-1} + \overline{\textbf{C}}\overline{\textbf{B}} u_k,~~~~~~y = \overline{\textbf{K}} * u
\end{align}

\begin{align}
\label{eq:conv_kernel_done}
    \overline{\textbf{K}} \in \mathbbm{R}^L := \mathcal{K}_L(\overline{\textbf{C}},\overline{\textbf{A}},\overline{\textbf{B}}) := \big(\overline{\textbf{C}}\overline{\textbf{A}}^{i}\overline{\textbf{B}}\big)_{i \in [L]} = \big(\overline{\textbf{C}}\overline{\textbf{B}}, \overline{\textbf{C}}\overline{\textbf{A}}\overline{\textbf{B}}, \dots, \overline{\textbf{C}}\overline{\textbf{A}}^{L-1}\overline{\textbf{B}} \big)
\end{align}
\endgroup

Equation (\ref{eq:conv_kernel}) is a non-circular convolutional kernel. \cite{gu2022efficiently} showed that under the condition that $\overline{\textbf{K}}$ is known, it could be solved very efficiently by a black-box Cauchy kernel computation pipeline. 

\subsection{Liquid Structural State-Space Models}
In this work, we construct a convolutional kernel corresponding to a linearized version of LTCs \citep{hasani2021liquid}; an expressive class of continuous-time neural networks that demonstrate attractive generalizability out-of-distribution and are dynamic causal models \citep{vorbach2021causal,friston2003dynamic,hasani2020natural}. In their general form, the state of a liquid time-constant network at each time-step is given by the set of ODEs described below \citep{hasani2021liquid}: 

\begingroup
\small
\begin{equation}
\label{eq:ltc}
    \frac{d\textbf{x}(t)}{dt} = - \underbrace{\Big[\bm{A} + \bm{B} \odot f(\textbf{x}(t),\textbf{u}(t), t, \theta)\Big]}_\text{Liquid time-constant} \odot \textbf{x}(t) + \bm{B} \odot f(\textbf{x}(t), \textbf{u}(t), t, \theta). 
\end{equation}
\endgroup

In this expression, $\textbf{x}^{(N \times 1)}(t)$ is the vector of hidden state of size $N$, $\textbf{u}^{(m \times 1)}(t)$ is an input signal with $m$ features, $\bm{A}^{(N \times 1)}$ is a time-constant state-transition mechanism, $\bm{B}^{(N \times 1)}$ is a bias vector, and $\odot$ represents the Hadamard product. $f(.)$ is a bounded nonlinearity parametrized by $\theta$. 

Our objective is to show how the liquid time-constant (i.e., an input-dependent state transition mechanism in state-space models can enhance its generalization capabilities by accounting for the covariance of the input samples. To do this, we linearize the LTC formulation of Eq. \ref{eq:ltc} in the following to better connect the model to SSMs. Let's dive in:

\noindent \textbf{Linear Liquid Time-Constant State-Space Model.} A Linear LTC SSM can be presented by the following coupled bilinear (first order bilinear Taylor approximation \citep{penny2005bilinear}) equation:

\begingroup
\small
\begin{align}
\label{eq:ltc_continuous}
    \dot{x}(t) = \big[\textbf{A} + \textbf{B}~u(t)]~x(t) + \textbf{B}~u(t),~~~~~~y(t) = \textbf{C}~x(t)     
\end{align}
\endgroup

Similar to (\ref{eq:ss_continuous}), $x(t)$ is an $N$-dimensional latent state, receiving a 1-dimensional input signal $u(t)$, and computing a 1-dimensional output signal $y(t)$. $\textbf{A}^{(N \times N)}$, $\textbf{B}^{(N \times 1)}$, and $\textbf{C}^{(1 \times N)}$. Note that \textbf{D} is set to zero for simplicity. In (\ref{eq:ltc_continuous}), the first $\textbf{B}~u(t)$ is added element-wise to $\textbf{A}$. This dynamical system allows the coefficient (state transition compartment) of state vector $x(t)$ to be input dependent which, as a result, allows us to realize more complex dynamics. 

\noindent \textbf{Discretization of Liquid-SSMs.} Similar to SSMs, Liquid-SSMs can also be discretized by a bilinear transform (trapezoidal rule) to construct a sequence-to-sequence model as follows:

\begingroup
\small
\begin{align}
\label{eq:liquid_ssm_discretized}
    x_k = \big( \overline{\textbf{A}} + \overline{\textbf{B}}~u_k\big)~x_{k-1} + \overline{\textbf{B}}~u_k,~~~~~~y_k = \overline{\textbf{C}}~x_k
\end{align}
\endgroup

The discretized parameters $\overline{\textbf{A}}$, $\overline{\textbf{B}}$, and $\overline{\textbf{C}}$ are identical to that of (\ref{eq:ss_disc_params}), which are function of the continuous-time coefficients $\textbf{A}$, $\textbf{B}$, and $\textbf{C}$, and the discretization step $\delta t$.

\noindent \textbf{Creating a Convolutional Representation of Liquid-SSMs.} Similar to (\ref{eq:unroll_ssm}), we first unroll the Liquid-SSM in time to construct a convolutional kernel of it. By assuming $x_{-1} = 0$, we have:

\begingroup
\footnotesize
\begin{align}
    \label{eq:unroll_liquid_ssm} \nonumber
     x_0 &= \overline{\textbf{B}} u_0,~~~~~y_0 = \overline{\textbf{C}}\overline{\textbf{B}} u_0 \\
     x_1 &= \overline{\textbf{A}}\overline{\textbf{B}} u_0 + \overline{\textbf{B}} u_1 {\color{violet} +~ \overline{\textbf{B}}^2 u_0 u_1},~~~~~y_1 = \overline{\textbf{C}}\overline{\textbf{A}}\overline{\textbf{B}} u_0 + \overline{\textbf{C}}\overline{\textbf{B}} u_1 {\color{violet} + \overline{\textbf{C}}\overline{\textbf{B}}^2 u_0 u_1} \\ \nonumber
     x_2 &= \overline{\textbf{A}}^2\overline{\textbf{B}} u_0 + \overline{\textbf{A}}\overline{\textbf{B}} u_1 + \overline{\textbf{B}} u_2 {\color{violet} +~ \overline{\textbf{A}}\overline{\textbf{B}}^2 u_0 u_1 +~ \overline{\textbf{A}}\overline{\textbf{B}}^2 u_0 u_2 +~ \overline{\textbf{B}}^2 u_1 u_2 +~ \overline{\textbf{B}}^3 u_0 u_1 u_2 } \\ \nonumber
     y_2 &= \overline{\textbf{C}}\overline{\textbf{A}}^2\overline{\textbf{B}} u_0 + \overline{\textbf{C}}\overline{\textbf{A}}\overline{\textbf{B}} u_1 + \overline{\textbf{C}}\overline{\textbf{B}} u_2 {\color{violet} +~ \overline{\textbf{C}}\overline{\textbf{A}}\overline{\textbf{B}}^2 u_0 u_1 +~ \overline{\textbf{C}}\overline{\textbf{A}}\overline{\textbf{B}}^2 u_0 u_2 +~ \overline{\textbf{C}}\overline{\textbf{B}}^2 u_1 u_2 +~ \overline{\textbf{C}}\overline{\textbf{B}}^3 u_0 u_1 u_2 },~~ \dots \nonumber
\end{align}
\endgroup

The resulting expressions of the Liquid-SSM at each time step consist of two types of weight configurations: 1. Weights corresponding to the mapping of individual time instances of inputs independently, shown in black in (\ref{eq:unroll_liquid_ssm}), and 2. Weights associated with all orders of auto-correlation of the input signal, shown in violet in (\ref{eq:unroll_liquid_ssm}). 
The first set of weights corresponds to the convolutional kernel of the simple SSM, shown by Eq. \ref{eq:conv_kernel} and Eq. \ref{eq:conv_kernel_done}, whereas the second set leads to the design of an additional input correlation kernel, which we call the \emph{liquid} kernel. These kernels generate the following input-output mapping:
\begingroup
\footnotesize
\begin{align}
\label{eq:conv_kernel_ltc} \nonumber
    y_k =~& \overline{\textbf{C}}\overline{\textbf{A}}^k\overline{\textbf{B}} u_0 + \overline{\textbf{C}}\overline{\textbf{A}}^{k-1}\overline{\textbf{B}} u_1 + \dots \overline{\textbf{C}}\overline{\textbf{A}}\overline{\textbf{B}} u_{k-1} + \overline{\textbf{C}}\overline{\textbf{B}} u_k ~+ \\ 
    & {\color{violet}
     \sum_{p=2}^\mathcal{P}  \forall \binom{k+1}{p} \text{~of~} u_i u_{i+1}~\dots~u_p~ \overline{\textbf{C}}\overline{\textbf{A}}^{(k+1-p-i)}\overline{\textbf{B}}^p u_i u_{i+1}~\dots~u_p} \\ \nonumber
   & \text{for} ~i \in \mathbbm{Z} \text{~and~} i \geq 0,~~~~~\rightarrow~~~~~~~~y =~ \overline{\textbf{K}} * u \nonumber +{\color{violet} \overline{\textbf{K}}_{\text{liquid}} * u_{\text{correlations}}}
\end{align}
\endgroup

For instance, let us assume we have a 1-dimensional input signal $u(t)$ of length $L=100$ on which we run the liquid-SSM kernel. We set the hyperparameters $\mathcal{P} = 4$. This value represents the maximum order of the correlation terms we would want to take into account to output a decision. This means that the signal $u_{\text{correlations}}$ in (\ref{eq:conv_kernel_ltc}) will contain all combinations of 2 order correlation signals $\binom{L+1}{2}$, $u_i u_j$, 3 order $\binom{L+1}{3}$, $u_i u_j u_k$ and 4 order signals $\binom{L+1}{4}$, $u_i u_j u_k u_l$. The kernel weights corresponding to this auto-correlation signal would be: 

\begin{align}
\overline{\textbf{K}}_{\text{liquid}} &* u_{\text{correlations}} = 
  \resizebox{.7\hsize}{!}{ $\Big[
        \overline{\textbf{C}}\overline{\textbf{A}}^{(k-1)}\overline{\textbf{B}}^2, 
        \dots,
        \overline{\textbf{C}}\overline{\textbf{B}}^2,
        \dots,
        \overline{\textbf{C}}\overline{\textbf{A}}^{(k-2)}\overline{\textbf{B}}^3,
        \dots,
        \overline{\textbf{C}}\overline{\textbf{B}}^3,
        \dots,
        \overline{\textbf{C}}\overline{\textbf{A}}^{(k-3)}\overline{\textbf{B}}^4,
        \dots,
        \overline{\textbf{C}}\overline{\textbf{B}}^4
\Big]$} *  \\ \nonumber
& \resizebox{.9\hsize}{!}{ $ \Big [
        u_0 u_1,
        \dots,
        u_{k-1}u_{k},
        \dots,
        u_0 u_1 u_2,
        \dots,
        u_{k-2} u_{k-1} u_{k},
        \dots,
        u_0 u_1 u_2 u_3,
        \dots,
       u_{k-3} u_{k-2} u_{k-1} u_{k} \Big ]^T
$}
\end{align}

Here, $u_{\text{correlations}}$ is a vector of length ${\binom{k+1}{2} + \binom{k+1 }{3} + \binom{k+1}{4}}$, and the kernel $\overline{\textbf{K}}_{\text{liquid}} \in \mathbbm{R}^{\binom{k+1}{2} + \binom{k+1}{3} + \binom{k+1}{4}}$. This additional kernel takes the temporal correlation of incoming input samples into consideration. This way Liquid-SSM give rise to a more general sequence modeling framework. The liquid convolutional kernel, $\overline{\textbf{K}}_{\text{liquid}}$ is as follows:

\begingroup
\small
\begin{align}
\label{eq:kb}
    \overline{\textbf{K}}_{\text{liquid}} \in \mathbbm{R}^{\Tilde{L}} := \mathcal{K}_L(\overline{\textbf{C}},\overline{\textbf{A}},\overline{\textbf{B}}) := \big(\overline{\textbf{C}}\overline{\textbf{A}}^{(\tilde{L}-i-p)}\overline{\textbf{B}}^p\big)_{i \in [\tilde{L}],~ p \in [\mathcal{P}]} = \big( \overline{\textbf{C}}\overline{\textbf{A}}^{\tilde{L}-2}\overline{\textbf{B}}^2, \dots, \overline{\textbf{C}}\overline{\textbf{B}}^p \big)
\end{align}
\endgroup

\begingroup
\small
\begin{algorithm}[t]
  \caption{\textsc{Liquid-S4 Kernel} - The S4 convolution kernel (highlighted in black) is used from \citet{gu2022efficiently} and \citet{gu2022parameterization}. Liquid kernel computation is highlighted in purple.}%
  \label{alg:liquid-s4-convolution}
  \begin{algorithmic}[1]
    \renewcommand{\algorithmicrequire}{\textbf{Input:}}
    \Require{S4 parameters \( \bm{\Lambda}, \bm{P}, \bm{B}, \bm{C} \in \mathbbm{C}^N \), step size \( \Delta \), {\color{violet}liquid kernel order $\mathcal{P}$, inputs seq length $L$, liquid kernel sequence length $\tilde{L}$}}
    \renewcommand{\algorithmicensure}{\textbf{Output:}}
    \Ensure{SSM convolution kernel \( \bm{\overline{K}} = \mathcal{K}_L(\bm{\overline{A}}, \bm{\overline{B}}, \bm{\overline{C}}) \) {\color{violet}and SSM liquid kernel \( \bm{\overline{K}}_{liquid} = \mathcal{K}_{\tilde{L}}(\bm{\overline{A}}, \bm{\overline{B}}, \bm{\overline{C}}) \)} for \( \bm{A} = \bm{\Lambda} - \bm{P}\bm{P}^* \)} (Eq. \ref{eq:conv_kernel_done})
    \State $\bm{\widetilde{C}} \gets \left(\bm{I} - \bm{\overline{A}}^L\right)^* \bm{\overline{C}}$
    \Comment{Truncate SSM generating function (SSMGF) to length \( L \)}
    \State
    \( \begin{bmatrix} k_{00}(\omega) & k_{01}(\omega) \\ k_{10}(\omega) & k_{11}(\omega) \end{bmatrix} \gets \left[ \bm{\widetilde{C}} \; \bm{P} \right]^* \left(\frac{2}{\Delta} \frac{1-\omega}{1+\omega} - \bm{\Lambda} \right)^{-1} \left[ \bm{B} \; \bm{P} \right] \)
    \label{step:cauchy}
    \Comment{Black-box Cauchy kernel}
    \State
    \( \bm{\hat{K}}(\omega) \gets \frac{2}{1+\omega} \left[ k_{00}(\omega) - k_{01}(\omega) (1 + k_{11}(\omega))^{-1} k_{10}(\omega) \right]  \)
    \Comment{Woodbury Identity} %
    \State \( \bm{\hat{K}} = \{\bm{\hat{K}}(\omega) : \omega = \exp(2\pi i \frac{k}{L})\} \)
    \Comment{Evaluate SSMGF at all roots of unity \( \omega \in \Omega_L \)}
    \State \( \bm{\overline{K}} \gets \mathsf{iFFT} (\bm{\hat{K}}) \)
    \Comment{Inverse Fourier Transform}
    \If{{\color{violet}Mode == KB}} \Comment{{\color{violet}Liquid-S4 Kernel as shown in Eq. \ref{eq:kb}}}
    \For{{\color{violet} $p$ in $\{2, \dots, \mathcal{P} \}$}}
    \State {\color{violet}\( \bm{\overline{K}}_{liquid=p} = \Big[ \bm{\overline{K}}_{(L-\tilde{L}, L)} \odot \bm{\overline{B}}^{p-1}_{(L-\tilde{L}, L)} \Big] * \textbf{J}_{\tilde{L}} \) \Comment{\(\textbf{J}_{\tilde{L}} \) is a backward identity matrix}}
    \State {\color{violet} \(\bm{\overline{K}}_{liquid}\text{.append}(\bm{\overline{K}}_{liquid=p}) \)}
    \EndFor
    \ElsIf{{\color{violet}Mode == PB}} \Comment{{\color{violet}Liquid-S4 Kernel of Eq. \ref{eq:kb} with $\bm{\overline{A}}$ reduced to Identity.}}
    \For{{\color{violet} $p$ in $\{2, \dots, \mathcal{P} \}$}}
    \State {\color{violet} \( \bm{\overline{K}}_{liquid=p} =  \bm{\overline{C}} \odot \bm{\overline{B}}^{p-1}_{(L-\tilde{L}, L)} \)}
    \State {\color{violet} \(\bm{\overline{K}}_{liquid}\text{.append}(\bm{\overline{K}}_{liquid=p}) \)}
    \EndFor
    \EndIf
  \end{algorithmic}
\end{algorithm}
\endgroup

\noindent \textbf{How to compute Liquid-S4 kernel efficiently?}
\citet{gu2022efficiently} showed that the S4 convolution kernel could be computed efficiently using the following elegant parameterization tricks:

\begin{itemize}
\item To obtain better representations in sequence modeling schemes by SSMs, instead of randomly initializing the transition matrix ${\textbf{A}}$, we can use the Normal Plus Low-Rank (NPLR) matrix below, called the Hippo Matrix \citep{gu2020hippo} which is obtained by the Scaled Legendre Measure (LegS) \citep{gu2021combining,gu2022efficiently}:
\begingroup
\small
\begin{equation}
  \label{eq:hippo}
  (\text{\textbf{HiPPO Matrix}})
  \qquad
  \bm{A}_{nk}
  =
  -
  \begin{cases}
    (2n+1)^{1/2}(2k+1)^{1/2} & \mbox{if } n > k \\
    n+1 & \mbox{if } n = k \\
    0 & \mbox{if } n < k
  \end{cases}
\end{equation}
\endgroup
\item The NPLR representation of this matrix is the following \citep{gu2022efficiently}:
\begin{equation}
    \label{eq:nplr}
    \bm{A} = \bm{V} \bm{\Lambda} \bm{V}^* - \bm{P} \bm{Q}^\top = \bm{V} \left( \bm{\Lambda} - \left(\bm{V}^* \bm{P}\right) (\bm{V}^*\bm{Q})^* \right) \bm{V}^*
  \end{equation}
Here, \( \bm{V} \in \mathbbm{C}^{N \times N} \) is a unitary matrix, \( \bm{\Lambda} \) is diagonal, and \( \bm{P}, \bm{Q} \in \mathbbm{R}^{N \times r} \) are the low-rank factorization. Eq. \ref{eq:hippo} is Normal plus low rank with r = 1 \citep{gu2022efficiently}. With the decomposition \ref{eq:nplr}, we can obtain $\textbf{A}$ over complex numbers in the form of Diagonal plus low-rank (DPLR) \citep{gu2022efficiently}.
\item Vectors $B_n$ and $P_n$ are initialized by $\bm{B_n} = (2n+1)^{\frac{1}{2}}$ and $\bm{P_n} = (n+1/2)^{\frac{1}{2}}$ \citep{gu2022parameterization}. Both vectors are trainable.
\item Furthermore, it was shown in \citet{gu2022parameterization} that with  Decomposition \ref{eq:nplr}, the eigenvalues of \textbf{A} might be on the right half of the complex plane, thus, result in numerical instability. To resolve this, \citet{gu2022parameterization} recently proposed to use the parametrization $ \bm{\Lambda} - \bm{P} \bm{P}^*$ instead of $\bm{\Lambda}  - \bm{P} \bm{P}^*$.
\item Computing the powers of $\bm{A}$ in direct calculation of the S4 kernel $\bm{\overline{K}}$ is computationally expensive. S4 computes the spectrum of $\bm {\overline{K}}$ instead of direct computations, which reduces the problem of matrix powers to matrix inverse computation \cite{gu2022efficiently}. S4 then computes this convolution kernel via a black-box Cauchy Kernel efficiently, and recovers $\bm{\overline{K}}$ by an inverse Fourier Transform (iFFT) \citep{gu2022efficiently}. 
\end{itemize}

$\overline{\textbf{K}}_{\text{liquid}}$ possess similar structure to the S4 kernel. In particular, we have:

\begin{proposition}
\label{prop:liquid kernel}
The liquid-S4 kernel for each order $p \in \mathcal{P}$, $\overline{\textbf{K}}_{\text{liquid}}$, can be computed by the anti-diagonal transformation (flip operation) of the product of the S4 convolution kernel, $\overline{\textbf{K}} = \big(\overline{\textbf{C}}\overline{\textbf{B}}, \overline{\textbf{C}}\overline{\textbf{A}}\overline{\textbf{B}}, \dots, \overline{\textbf{C}}\overline{\textbf{A}}^{L-1}\overline{\textbf{B}} \big)$, and a vector $\bm{\overline{B}}^{p-1} \in \mathbbm{R}^N$.
\end{proposition}

The proof is given in Appendix. Proposition \ref{prop:liquid kernel} indicates that the liquid-s4 kernel can be obtained from the precomputed S4 kernel and a Hadamard product of that kernel with the transition vector $\bm{\overline{B}}$ powered by the chosen liquid order. This is illustrated in Algorithm \ref{alg:liquid-s4-convolution}, lines 6 to 10, corresponding to a mode we call KB, which stands for Kernel $\times$ B.

Additionally, we introduce a simplified Liquid-S4 kernel that is easier to compute while being as expressive as or even better performing than the KB kernel. To obtain this, we set the transition matrix $\bm{\overline{A}}$ in Liquid-S4 of Eq. \ref{eq:kb}, with an identity matrix, only for the input correlation terms. This way, the liquid-s4 Kernel for a given liquid order $p \in \mathcal{P}$ reduces to the following expression: 

\begin{align}
\label{eq:polyb}
(\text{\textbf{Liquid-S4 - PB}})
  \qquad
    \overline{\textbf{K}}_{\text{liquid}=p} \in \mathbbm{R}^{\Tilde{L}} := \mathcal{K}_L(\overline{\textbf{C}},\overline{\textbf{B}}) := \big(\overline{\textbf{C}}\overline{\textbf{B}}^p\big)_{i \in [\tilde{L}],~ p \in [\mathcal{P}]} %= \big( \overline{\textbf{C}}\overline{\textbf{B}}^p, \dots, \overline{\textbf{C}}\overline{\textbf{B}}^p \big)
\end{align}

We call this kernel Liquid-S4 - PB, as it is obtained by powers of the vector $\bm{\overline{B}}$. The computational steps to get this kernel is outlined in Algorithm \ref{alg:liquid-s4-convolution} lines 11 to 15.

\opexpr{(62.75 + 89.02 + 91.20 + 89.50 + 94.8 + 96.66)}{b}
\opdiv*[maxdivstep=4]{b}{6}{q}{r}
 \opfloor{q}{2}{a}

\begin{table}[t] 
    \centering
    \caption{Performance on Long Range Arena Tasks. Numbers indicate validation accuracy (standard deviation). The accuracy of models denoted by * are reported from \citep{tay2020long}. Methods denoted by ** are reported from \citep{gu2022efficiently}. The rest of the models' performance results are reported from the cited paper. Liquid-S4 is used with its PB kernel.}
    \begin{adjustbox}{width=1\columnwidth}
    \begin{tabular}{lcccccc|c}
    \toprule
        Model & ListOps & IMDB & AAN	& CIFAR	& Pathfinder & Path-X & Avg. \\
        (input length) & 2048 & 2048 & 4000 & 1024 & 1024 & 16384 & \\
        \midrule
        Random$^{*}$ & 10.00 & 50.00 & 50.00 & 10.00 & 50.00 & 50.00 & 36.67 \\
        Transformer$^{*}$ \citep{vaswani2017attention} & 36.37 & 64.27 & 57.46 & 42.44 & 71.40 & x & 54.39 \\
        Local Att.$^{*}$ \citep{tay2020long} & 15.82 &	52.98 &	53.39 &	41.46 &	66.63 &	x &	46.06 \\
        Sparse Transformer$^{*}$ \citep{child2019generating} & 17.07	& 63.58 &	59.59 &	44.24 &	71.71	& x	& 51.24 \\
        Longformer$^{*}$ \citep{beltagy2020longformer}  & 35.63	& 62.85	& 56.89	& 42.22	& 69.71	& x & 53.46 \\
        Linformer$^{*}$ \citep{wang2020linformer} & 16.13	& 65.90	& 53.09	& 42.34	& 75.30	& x & 	50.55 \\
Reformer$^{*}$ \citep{kitaev2019reformer} & 37.27 & 56.10 & 53.40 & 38.07 & 68.50 & x & 50.56 \\
Sinkhorn Trans.$^{*}$ \citep{tay2020sparse} & 33.67 & 61.20 & 53.83 & 41.23 & 67.45 & x & 51.23 \\
BigBird$^{*}$ \citep{zaheer2020big} & 36.05	& 64.02	& 59.29	& 40.83	& 74.87	& x &	55.01 \\
Linear Trans.$^{*}$ \citep{katharopoulos2020transformers}& 16.13 & 65.90 & 53.09 & 42.34 & 75.30 & x & 50.46 \\
Performer$^{*}$  \citep{choromanski2020rethinking} & 18.01 & 65.40 & 53.82 & 42.77 & 77.05 & x & 51.18 \\
\midrule
FNet$^{**}$ \citep{lee2021fnet} & 35.33 & 65.11 & 59.61 & 38.67 & 77.80 & x & 54.42 \\
Nyströmformer$^{**}$ \citep{xiong2021nystromformer} & 37.15 & 65.52 & 79.56 & 41.58 & 70.94 & x &  57.46 \\
Luna-256$^{**}$ \cite{ma2021luna} & 37.25 & 64.57 & 79.29 & 47.38 & 77.72 & x & 59.37 \\
H-Transformer-1D$^{**}$ \citep{zhu2021h} & 49.53 & 78.69 & 63.99 & 46.05 & 68.78 & x & 61.41 \\
\midrule
CDIL \citep{cheng2022classification} & 44.05 &  86.78 &  85.36 & 66.91 & 91.70 & x & 74.96\\
\midrule
DSS \citep{gupta2022diagonal} & 57.6 & 76.6 & 87.6 & 85.8 & 84.1 & 85.0 & 79.45 \\
S4 (original)$^{**}$ \citep{gu2022efficiently} & 58.35 & 76.02 & 87.09 & 87.26 & 86.05 & 88.10 & 80.48 \\
%S4 v2$^{**}$ \citep{gu2022efficiently} & 59.5	& 86.5 & 91.0 &	88.5 &	94.0 &	96.0  & 85.91 \\
S4-LegS \citep{gu2022parameterization} & 59.60 (0.07)  & 86.82 (0.13) & 90.90 (0.15)  & 88.65 (0.23)  & 94.20 (0.25)  & \underline{96.35} & 86.09 \\
S4-FouT \citep{gu2022parameterization} & 57.88 (1.90) & 86.34 (0.31)  & 89.66 (0.88) &  89.07 (0.19)  & \underline{94.46} (0.26) & x & 77.90 \\
S4-LegS/FouT \citep{gu2022train} & 60.45 (0.75) & 86.78 (0.26)  & 90.30 (0.28)  & 89.00 (0.26)  & \underline{94.44} (0.08)  &  x & 78.50 \\
S4D-LegS \citep{gu2022parameterization} & 60.47 (0.34)  &  86.18 (0.43) & 89.46 (0.14)  & 88.19 (0.26)  & 93.06 (1.24)  & 91.95 & 84.89 \\
S4D-Inv \citep{gu2022parameterization} & 60.18 (0.35)  & 87.34 (0.20)  & \underline{91.09} (0.01)  & 87.83 (0.37) & 93.78 (0.25) &  92.80 & 85.50 \\
S4D-Lin \citep{gu2022parameterization} & 60.52 (0.51)  & 86.97 (0.23)  & 90.96 (0.09)  & 87.93 (0.34)  & 93.96 (0.60)  & x & 78.39 \\
S5 \citep{smith2022simplified} & 61.00 & 86.51 & 88.26 & 86.14 & 87.57 & 85.25 & 82.46 \\
\midrule
\textbf{Liquid-S4} (ours) & \textbf{62.75} (0.2) & \textbf{89.02} (0.04) & \textbf{91.20} (0.01) & \textbf{89.50} (0.4) & \textbf{94.8} (0.2) & \textbf{96.66}(0.001) & \textbf{\opprint{a}} \\
& p = 5 & p=6  & p=2 & p=3 & p=2 & p=2 & \\
         \bottomrule
    \end{tabular}
    \end{adjustbox}
    \label{tab:lra}
\end{table}

\noindent \textbf{Computational Complexity of the Liquid-S4 Kernel.} The computational complexity of the S4-Legs Convolutional kernel solved via the Cauchy Kernel is $\mathcal{\tilde{O}}(N + L)$, where N is the state-size, and L is the sequence length [\citet{gu2022efficiently}, Theorem 3]. Liquid-S4 both in KB and PB modes can be computed in $\mathcal{\tilde{O}}(N + L + p_{max} \tilde{L})$. The added time complexity in practice is tractable. This is because we usually select the liquid orders, $p$, to be less than 10 (typically $p_{max}=3$, and $\tilde{L}$ which is the number of terms we use to compute the input correlation vector, $u_{correlation}$, is typically two orders of magnitude smaller than the seq length. 

% \section{Liquid-S4 kernel approximations}
% Computing the entire Liquid-S4 kernel may not be feasible for some tasks with very long sequence lengths or very large models due to the increased computation and memory requirement.
% In such cases, we can approximate the kernel by computing only a subset of it, which reduces the memory requirements while still providing some benefits of the liquid model.
% In particular, we propose to compute the terms directly 
% \begin{equation}
%     \overline{\textbf{C}}\overline{\textbf{B}}^k u_i u_{i_1} \dots u_{i+k-1},
% \end{equation}
% up to a given order $k$. As the terms $u_i u_{i_1} \dots u_{i+k-1}$ appear consecutive, their product can be computed via simple padding operations and $k$ vector-vector products. 
% We name this kernel approximation "polyb" kernel because it only considers the polynomials of $\overline{\textbf{B}}$

\section{Experiments with Liquid-S4} 

In this section, we present an extensive evaluation of Liquid-S4 on sequence modeling tasks with very long-term dependencies and compare its performance to a large series of baselines ranging from advanced Transformers and Convolutional networks to many variants of State-space models. In the following, we first outline the baseline models we compare against. We then list the datasets we evaluated these models on and finally present results and discussions. 

\noindent \textbf{Baselines.} We consider a broad range of advanced models to compare liquid-S4 with. These baselines include transformer variants such as vanilla Transformer \citep{vaswani2017attention}, Sparse Transformers \citep{child2019generating}, a Transformer model with local attention \citep{tay2020long}, Longformer \citep{beltagy2020longformer}, Linformer \citep{wang2020linformer}, Reformer \citep{kitaev2019reformer}, Sinkhorn Transformer \citep{tay2020sparse}, BigBird \citep{zaheer2020big}, Linear Transformer \citep{katharopoulos2020transformers}, and Performer \citep{choromanski2020rethinking}. We also include architectures such as FNets \citep{lee2021fnet}, Nystro\"mformer \citep{xiong2021nystromformer}, Luna-256 \citep{ma2021luna}, H-Transformer-1D \citep{zhu2021h}, and Circular Diluted Convolutional neural networks (CDIL) \citep{cheng2022classification}. We then include a full series of state-space models and their variants such as diagonal SSMs (DSS) \citep{gupta2022diagonal}, S4 \citep{gu2022efficiently}, S4-legS, S4-FouT, S4-LegS/FouT \citep{gu2022train}, S4D-LegS \citep{gu2022parameterization}, S4D-Inv, S4D-Lin and the Simplified Structural State-space models (S5) \citep{smith2022simplified}.

\begin{table}[t]
    \centering
    \caption{Performance on BIDMC Vital Signs dataset. Numbers indicate RMSE on the test set. The accuracy of models denoted by * is reported from \citep{gu2022parameterization}. The rest of the models' performance results are reported from the cited paper.}
    \begin{adjustbox}{width=0.65\columnwidth}
    \begin{tabular}{lccc}
    \toprule
     & &  \textbf{BIDMC} & \\
     \toprule
        Model & HR & RR & SPO2 \\
        \midrule
UnICORNN \citep{rusch2021coupled} & 1.39 & 1.06 & 0.869 \\
coRNN \citep{rusch2021coupled} & 1.81 & 1.45 & - \\
CKConv$^{*}$ & 2.05 & 1.214 & 1.051 \\
NRDE \citep{morrill2021neural} & 2.97 & 1.49 & 1.29 \\
LSTM$^{*}$ & 10.7 & 2.28 & - \\
Transformer$^{*}$ & 12.2 & 2.61 & 3.02 \\
XGBoost \citep{tan2021time} & 4.72 & 1.67 & 1.52 \\
Random Forest \citep{tan2021time} & 5.69 & 1.85 & 1.74 \\
Ridge Regress. \citep{tan2021time} &  17.3 & 3.86 & 4.16 \\
\midrule
        S4-LegS$^{*}$ \citep{gu2022parameterization} & 0.332 (0.013) & 0.247 (0.062) & 0.090 (0.006) \\ 
        S4-FouT$^{*}$ \citep{gu2022parameterization} & 0.339 (0.020) & 0.301 (0.030) & \underline {0.068} (0.003) \\ 
        S4D-LegS$^{*}$ \citep{gu2022parameterization} & 0.367 (0.001) & 0.248 (0.036) & 0.102 (0.001) \\ 
        S4-(LegS/FouT)$^{*}$ \citep{gu2022parameterization} & 0.344 (0.032) & \underline{0.163} (0.008) & 0.080 (0.007) \\
        S4D-Inv$^{*}$ \citep{gu2022parameterization} & 0.373 (0.024) & 0.254 (0.022) & 0.110 (0.001) \\ 
        S4D-Lin$^{*}$ \citep{gu2022parameterization} & 0.379 (0.006) & 0.226 (0.008) & 0.114 (0.003) \\ 
        \midrule
        \textbf{Liquid-S4} (ours) & \textbf{0.303} (0.002) & \textbf{0.158} (0.001) & \textbf{0.066} (0.002)\\
        & p=3 & p=2 & p=4 \\
        \bottomrule
    \end{tabular}
    \end{adjustbox}
    \label{tab:bidmc}
\end{table}

\noindent\textbf{Datasets.} We first evaluate Liquid-S4's performance on the well-studied \textbf{Long Range Arena (LRA)} benchmark \citep{tay2020long}, where Liquid-S4 outperforms other S4 and S4D variants in every task pushing the state-of-the-art further with an average accuracy of \textbf{87.32\%}. LRA dataset includes six tasks with sequence lengths ranging from 1k to 16k. 

We then report Liquid-S4's performance compared to other S4, and S4D variants as well as other models, on the \textbf{BIDMC Vital Signals} dataset \citep{pimentel2016toward,goldberger2000physiobank}. BIDMC uses bio-marker signals of length 4000 to predict Heart rate (HR), respiratory rate (RR), and blood oxygen saturation (SpO2). 

We also experiment with the \textbf{sCIFAR dataset} that consists of the classification of flattened images in the form of 1024-long sequences into 10 classes.

Finally, we perform \textbf{RAW Speech Command (SC) recognition with FULL 35 LABELS} as conducted very recently in the updated S4 article \citep{gu2022efficiently}.\footnote{It is essential to denote that there is a modified speech command dataset that restricted the dataset to only 10 output classes and is used in a couple of works (see for example \protect\citep{kidger2020neural,gu2021combining,romero2021ckconv,romero2021flexconv}). Aligned with the updated results reported in \citep{gu2022efficiently} and \citep{gu2022parameterization}, we choose not to break down this dataset and use the full-sized benchmark.} SC dataset contains sequences of length 16k to be classified into 35 commands. \citet{gu2022efficiently} introduced a new test case setting to assess the performance of models (trained on 16kHz sequences) on sequences of length 8kHz. S4 and S4D perform exceptionally well in this zero-shot test scenario.

\begin{figure}[t]
    \centering
    \includegraphics[width=0.65\textwidth]{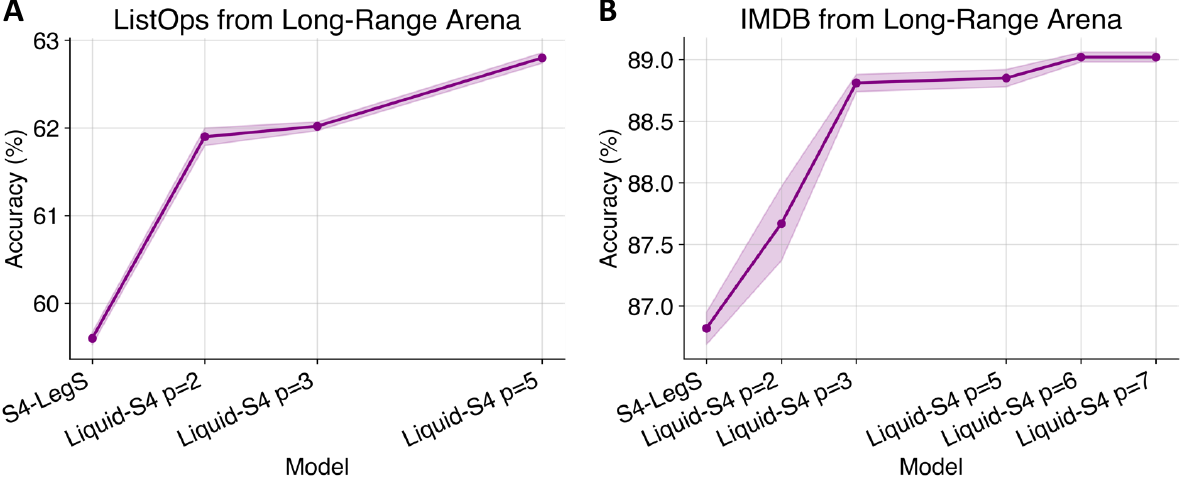}
    \caption{Performance vs Liquid Order in Liquid-S4 for A) ListOps, and B) IMDB datasets. (n=3)}
    \label{fig:liquid_order}
\end{figure}

\subsection{Results on Long Range Arena}

Table \ref{tab:lra} depicts a comprehensive list of baselines benchmarked against each other on six long-range sequence modeling tasks in LRA. We observe that Liquid-S4 instances (all use the PB kernel with a scaled Legendre (LegS) configuration) with a small liquid order, $p$, ranging from 2 to 6, consistently outperform all baselines in all six tasks, establishing the new SOTA on LRA with an average performance of \textbf{87.32\%}. In particular, on ListOps, Liquid-S4 improves S4-LegS performance by more than 3\%, on character-level IMDB by 2.2\%, and on 1-D pixel-level classification (CIFAR) by 0.65\%, while establishing the-state-of the-art on the hardest LRA task by gaining \textbf{96.54\%} accuracy. Liquid-S4 performs on par with improved S4 and S4D instances on both AAN and Pathfinder tasks. 

\begin{table}[t]
    \centering
    \caption{Performance on sCIFAR dataset. Numbers indicate Accuracy (standard deviation). The accuracy of baseline models is reported from Table 9 of \citet{gu2022parameterization}.}
    \begin{adjustbox}{width=0.45\textwidth}
    \begin{tabular}{lccc}
\toprule
Model & Accuracy \\
\midrule
Transformer \citep{trinh2018learning} & 62.2 \\
FlexConv \citep{romero2021flexconv} & 80.82 \\
TrellisNet \citep{bai2018trellis} & 73.42 \\
LSTM \citep{hochreiter1997long} & 63.01 \\
r-LSTM \citep{trinh2018learning} & 72.2 \\
UR-GRU \citep{gu2020improving} & 74.4 \\
HiPPO-RNN \citep{gu2020hippo} & 61.1 \\
LipschitzRNN \citep{erichson2021lipschitz} & 64.2 \\
\midrule
S4-LegS \citep{gu2022parameterization} & \underline{91.80} (0.43) \\
S4-FouT \citep{gu2022parameterization} & 91.22 (0.25) \\
S4-(LegS/FouT) \citep{gu2022parameterization} & 91.58 (0.17) \\
\midrule
S4D-LegS \citep{gu2022parameterization} & 89.92 (1.69) \\
S4D-Inv \citep{gu2022parameterization} & 90.69 (0.06) \\
S4D-Lin \citep{gu2022parameterization} & 90.42 (0.03) \\
S5 \cite{smith2022simplified} &  89.66 \\
\midrule
Liquid-S4 (ours) & \textbf{92.02} (0.14) \\
& p=3\\
\bottomrule
    \end{tabular}
    \end{adjustbox}
    \label{tab:scifar}
\end{table}

The performance of SSM models is generally well-beyond what advanced Transformers, RNNs, and Convolutional networks achieve on LRA tasks, with the Liquid-S4 variants standing on top. It is worth noting that Liquid-S4 kernels perform better with smaller kernel sizes (See more details on this in Appendix); For instance, on ListOps and IMDB, their individual liquid-S4 kernel state-size could be as small as seven units. This significantly reduces the parameter count in Liquid-S4 in comparison to other variants.

\noindent\textbf{The impact of increasing Liquid Order $p$.} Figure \ref{fig:liquid_order} illustrates how increasing the liquid order, $p$, can consistently improve performance on ListOps and IMDB tasks from LRA. 

\subsection{Results on BIDMC Vital Signs}

\begin{table}[t] 
    \centering
    \caption{Performance on RAW Speech Command dataset with \textbf{FULL 35 Labels} and with the reduced ten classes.%\footnote{It is essential to denote that there is a modified speech command dataset restricted the dataset to only ten output classes, and is used in a couple of works (see for example \protect\citet{kidger2020neural,gu2021combining,romero2021ckconv,romero2021flexconv}). Aligned with the updated results reported in \citep{gu2022efficiently} and \citep{gu2022parameterization}, we choose not to break down this dataset and use the full-sized benchmark.} 
     Numbers indicate validation accuracy. The accuracy of baseline models is reported from Table 11 of \citep{gu2022parameterization}.}
\begin{adjustbox}{width=0.65\columnwidth}
\begin{tabular}{lccc}
\toprule
& &  \multicolumn{2}{c}{\textbf{SC FULL Labels}} \\
\midrule
Model & Parameters & 16kHz & 8kHz \\
\midrule
InceptionNet \citep{nonaka2021depth} & 481K & 61.24 (0.69) & 05.18 (0.07) \\
ResNet-18 & 216K & 77.86 (0.24) & 08.74 (0.57) \\
XResNet-50 & 904K & 83.01 (0.48) & 07.72 (0.39) \\
ConvNet & 26.2M & 95.51 (0.18) & 07.26 (0.79) \\
\midrule
S4-LegS \citep{gu2022parameterization} & 307K & 96.08 (0.15) & 91.32 (0.17) \\
S4-FouT \citep{gu2022parameterization} & 307K & 95.27 (0.20) & 91.59 (0.23) \\
S4-(LegS/FouT) \citep{gu2022parameterization} & 307K & 95.32 (0.10) & 90.72 (0.68) \\
\midrule
S4D-LegS \citep{gu2022parameterization} & 306K & 95.83 (0.14) & 91.08 (0.16) \\
S4D-Inv \citep{gu2022parameterization} & 306K & \underline{96.18} (0.27) & \textbf{91.80} (0.24) \\
S4D-Lin \citep{gu2022parameterization} & 306K & \underline{96.25} (0.03) & \underline{91.58} (0.33) \\
\midrule
Liquid-S4 (ours) & \textbf{224K} & \textbf{96.78} (0.05) & 90.00 (0.25)  \\
& & p=2 & p=2 \\
\bottomrule
\end{tabular}
\end{adjustbox}
\label{tab:sc}
\end{table}
Table \ref{tab:bidmc} demonstrates the performance of a variety of classical and advanced baseline models on the BIDMC dataset for all three heart rate (HR), respiratory rate (RR), and blood oxygen saturation (SpO2) level prediction tasks. We observe that Liquid-s4 with a PB kernel of order $p=3$, $p=2$, and $p=4$, perform better than all S4 and S4D variants. It is worth denoting that Liquid-S4 is built by the same parametrization as S4-LegS (which is the official S4 model reported in the updated S4 report \citep{gu2022efficiently}). In RR, Liquid-S4 outperforms S4-LegS by a significant margin of 36\%. On SpO2, Liquid-S4 performs 26.67\% better than S4-Legs. On HR, Liquid-S4 outperforms S4-Legs by 8.7\% improvement in performance. The hyperparameters are given in Appendix.

\subsection{Results on 1-D Pixel-level Image Classification}
Similar to the previous tasks, a Liquid-S4 network with PB kernel of order $p=3$ outperforms all variants of S4 and S4D while being significantly better than Transformer and RNN baselines as summarized in Table \ref{tab:scifar}. The hyperparameters are given in Appendix.

\subsection{Results on Speech Commands}
Table \ref{tab:sc} demonstrates that Liquid-S4 with liquid order $p=2$ achieves the best performance amongst all benchmarks on the 16KHz testbed with full dataset. Liquid-S4 also performs competitively on the half-frequency zero-shot experiment, while it does not realize the best performance. Although the task is solved to a great degree, the reason could be that liquid kernel accounts for covariance terms. This might influence the learned representations in a way that hurts performance by a small margin in this zero-shot experiment. The hyperparameters are given in Appendix.

It is essential to denote that there is a modified speech command dataset that restricts the dataset to only ten output classes, namely SC10, and is used in a couple of works (see for example \protect\citep{kidger2020neural,gu2021combining,romero2021ckconv,romero2021flexconv}). Aligned with the updated results reported in \citep{gu2022efficiently} and \citep{gu2022parameterization}, we choose not to break down this dataset and report the full-sized benchmark in the main paper. Nevertheless, we conducted an experiment with SC10 and showed that even on the reduced dataset, with the same hyperparameters, we solved the task with a SOTA accuracy of 98.51\%. The results are presented in Table \ref{tab:sc_1}.

\section{Conclusions}
We showed that structural state-space models could be considerably improved in performance if they are formulated by a linear liquid time-constant kernel, namely Liquid-S4. Liquid-S4 kernels are obtainable with minimal effort with their kernel computing the similarities between time-lags of the input signals in addition to the main S4 diagonal plus low-rank parametrization. Liquid-S4 kernels with Smaller parameter counts achieve SOTA performance on all six tasks of the Long-range arena dataset, on BIDMC heart rate, respiratory rate, and blood oxygen saturation, on sequential 1-D pixel-level image classification, and on Speech command recognition. 

\section*{Acknowledgments}
This work is supported by The Boeing Company and the Office of Naval Research (ONR) Grant N00014-18-1-2830.

% \bibliographystyle{abbrvnat}
% \bibliography{references}

\clearpage
\beginsupplement

\section*{Supplementary Materials}

\section{Proof of Proposition \ref{prop:liquid kernel}}
\begin{proof}
This can be shown by unrolling the S4 convolution kernel and multiplying its components with $\bm{\overline{B}}^{p-1}$, performing an anti-diagonal transformation to obtain the corresponding liquid S4 kernel:
\begin{equation} \nonumber
\overline{\textbf{K}} = \big(\overline{\textbf{C}}\overline{\textbf{B}}, \overline{\textbf{C}}\overline{\textbf{A}}\overline{\textbf{B}}, \overline{\textbf{C}}\overline{\textbf{A}}^2\overline{\textbf{B}}, \dots, \overline{\textbf{C}}\overline{\textbf{A}}^{L-1}\overline{\textbf{B}} \big)
\end{equation}
For $p=2$ (correlations of order 2), S4 kernel should be multiplied by $\bm{\overline{B}}$. The resulting kernel would be:

\begin{equation} \nonumber
\big(\overline{\textbf{C}}\overline{\textbf{B}}^2, \overline{\textbf{C}}\overline{\textbf{A}}\overline{\textbf{B}}^2, \overline{\textbf{C}}\overline{\textbf{A}}^2\overline{\textbf{B}}^2, \dots, \overline{\textbf{C}}\overline{\textbf{A}}^{L-1}\overline{\textbf{B}}^2 \big)
\end{equation}

\noindent We obtain the liquid kernel by flipping the above kernel to be convolved with the 2-term correlation terms (p=2): 

\begin{equation} \nonumber
    \overline{\textbf{K}}_{\text{liquid=2}} = \big(\overline{\textbf{C}}\overline{\textbf{A}}^{L-1}\overline{\textbf{B}}^2, \dots, \overline{\textbf{C}}\overline{\textbf{A}}^2\overline{\textbf{B}}^2, \overline{\textbf{C}}\overline{\textbf{A}}\overline{\textbf{B}}^2, \overline{\textbf{C}}\overline{\textbf{B}}^2 \big)
\end{equation}

\noindent Similarly, we can obtain liquid kernels for higher liquid orders and obtain the statement of the proposition.
\end{proof}

\section{Hyperparameters}

\textbf{Learning Rate.} Liquid-S4 generally requires a smaller learning rate compared to S4 and S4D blocks.

\noindent \textbf{Setting $\Delta t_{max}$ and $\Delta t_{min}$} We set $\Delta t_{max}$ for all experiments to 0.2, while the $\Delta t_{min}$ was set based on the recommendations provided in \citep{gu2022train} to be proportional to $\propto \frac{1}{\text{seq length}}$.

\noindent \textbf{Causal Modeling vs. Bidirectional Modeling} Liquid-S4 works better when it is used as a causal model, i.e., with no bidirectional configuration.

\noindent \textbf{$d_state$} We observed that liquid-S4 PB kernel performs best with smaller individual state sizes $d_state$. For instance, we achieve SOTA results in ListOps, IMDB, and Speech Commands by a state size set to 7, significantly reducing the number of required parameters to solve these tasks.

\noindent \textbf{Choice of Liquid-S4 Kernel} In all experiments, we choose our simplified PB kernel over the KB kernel due to the computational costs and performance. We recommend the use of PB kernel. 

\noindent  \textbf{Choice of parameter $p$ in liquid kernel.} In all experiments, start off by setting $p$ or the liquidity order to 2. This means that the liquid kernel is going to be computed only for correlation terms of order 2. In principle, we observe that higher $p$ values consistently enhance the representation learning capacity of liquid-S4 modules, as we showed in all experiments. We recommend $p=3$ as a norm to perform experiments with Liquid-S4.

\noindent The kernel computation pipeline uses the PyKeops package \citep{JMLR:v22:20-275} for large tensor computations without memory overflow. 

\noindent All reported results are validation accuracy (similar to \citet{gu2022efficiently}) performed with 2 to 3 different random seeds, except for the BIDMC dataset, which reports accuracy on the test set.

\begin{table}[H]
    \centering
    \caption{Hyperparameters for obtaining best performing models. BN= Batch Normalization, LN = Layer normalization, WD= Weight decay.}
    \begin{adjustbox}{width=1\columnwidth}
    \begin{tabular}{lllllllllll}
        \toprule
               & \textbf{Depth} & \textbf{Features} $H$  & \textbf{State Size} & \textbf{Norm} & \textbf{Pre-norm} & \textbf{Dropout} & \textbf{LR} & \textbf{Batch Size} & \textbf{Epochs} & \textbf{WD} \\
        \midrule
        \textbf{ListOps} & 9 & 128 & 7 & BN & True & 0.01 & 0.002 & 12 & 30 & 0.03 \\
        \textbf{Text (IMDB)} & 4 & 128 & 7 & BN & True & 0.1 & 0.003 & 8 & 50 & 0.01 \\
        \textbf{Retrieval (AAN)} & 6 & 256 & 64 & BN & False & 0.2 & 0.005 & 16 & 20 & 0.05\\
        \textbf{Image (CIFAR)} & 6 & 512 & 512 & LN & False & 0.1 & 0.01 & 16 & 200 & 0.03 \\
        \textbf{Pathfinder} & 6 & 256 & 64 & BN & True & 0.0 & 0.0004 & 4 & 200 & 0.03 \\
        \textbf{Path-X} & 6 & 320 & 64 & BN & True & 0.0 & 0.001 & 8 & 60 & 0.05 \\
        \midrule
        \textbf{Speech Commands} & 6 & 128 & 7 & BN & True & 0.0  & 0.008 & 10 & 50 & 0.05\\
        \midrule
        \textbf{BICMD (HR)} & 6 & 128 & 256 & LN & True & 0.0 & 0.005 & 32 & 500 & 0.01\\
        \textbf{BICMD (RR)} & 6 & 128 & 256 & LN & True & 0.0 & 0.01 & 32 & 500 & 0.01 \\
        \textbf{BICMD (SpO2)} & 6 & 128 & 256 & LN & True & 0.0 & 0.01 & 32 & 500 & 0.01 \\
        \midrule
        \textbf{sCIFAR} & 6 & 512 & 512 & LN & False & 0.1 & 0.01 & 50 & 200 & 0.03 \\
        \bottomrule
    \end{tabular}
    \end{adjustbox}
    \label{tab:hyperparams}
\end{table}

% \begin{table}[t] 
%     \centering
%     \caption{Liquid-S4 Ablation -- on Long Range Arena Tasks. Numbers indicate validation accuracy (standard deviation). Liquid networks ablation.}
%     \begin{adjustbox}{width=1\columnwidth}
%     \begin{tabular}{lcccccc}
%     \toprule
%         Model & ListOps & IMDB & AAN	& CIFAR	& Pathfinder & Path-X \\
%         (input length) & 2048 & 2048 & 4000 & 1024 & 1024 & 16384 \\
%         \midrule
% S4-LegS \citep{gu2022parameterization} & 59.60  & 86.82 & 90.90  & 88.65  & 94.20   & 96.35 \\
% Liquid-S4 KB p=2  & 61.50  & 87.95 & --  & -- & -- & --  \\ 
% Liquid-S4 PB p=2 & 61.90  & 87.67 & 91.10  & 89.30 & 94.5 & 96.00  \\ 
% Liquid-S4 PB p=3 & 62.02  & 88.81 & --  & -- & -- & --  \\ 
% Liquid-S4 PB p=5 & 62.80  & 88.85 & --  & -- & -- & --  \\ 
% \bottomrule
%     \end{tabular}
%     \end{adjustbox}
%     \label{tab:lra_more_liquids}
% \end{table}

\begin{table}[H] 
    \centering
    \caption{Performance on RAW Speech Command dataset with the reduced ten classes (SC10) dataset.%\footnote{It is essential to denote that there is a modified speech command dataset that restricted the dataset to only 10 output classes and is used in a couple of works (see for example \protect\citet{kidger2020neural,gu2021combining,romero2021ckconv,romero2021flexconv}). Aligned with the updated results reported in \citep{gu2022efficiently} and \citep{gu2022parameterization}, we choose not to break down this dataset and use the full-sized benchmark.} 
     Numbers indicate validation accuracy. The accuracy of baseline models is reported from Table 5 of \citep{gu2022efficiently}. x stands for infeasible computation on a single GPU or not applicable as stated in Table 10 of \citep{gu2022efficiently}. The hyperparameters for Liquid-S4 are the same as the ones reported for Speech Commands Full Dataset reported in Table \ref{tab:hyperparams}.}
\begin{adjustbox}{width=0.4\columnwidth}
\begin{tabular}{lll}
\toprule
& \multicolumn{2}{c}{\textbf{SC10}} \\
\midrule
Model & 16kHz & 8kHz \\
\midrule
Transformer & x & x \\
Performer & 30.77 & 30.68 \\
ODE-RNN & x & x \\
NRDE & 16.49 & 15.12 \\
ExpRNN & 11.6 & 10.8 \\
LipschitzRNN & x & x \\
CKConv & 71.66 & 65.96 \\
WaveGAN-D &  96.25 & x \\
LSSL \citep{gu2021combining} & x & x \\
S4-LegS \citep{gu2022efficiently} & \text98.32 & \textbf{96.30} \\
\midrule
Liquid-S4 (ours) &  \textbf{98.51} & 95.9 \\
& p=2 & p=2 \\
\bottomrule
\end{tabular}
\end{adjustbox}
\label{tab:sc_1}
\end{table}

\end{document}